\documentclass[10pt,letterpaper]{article}

\usepackage[letterpaper,margin=0.72in,columnsep=0.24in]{geometry}
\usepackage{booktabs}
\usepackage{array}
\usepackage{url}
\usepackage{xurl}
\usepackage[numbers,sort&compress]{natbib}
\usepackage[colorlinks=true,linkcolor=black,citecolor=black,urlcolor=black,breaklinks=true]{hyperref}
\usepackage{tikz}
\usetikzlibrary{arrows.meta,positioning}

\title{\bfseries When Aggregate Alignment Misleads: Auditing Policy Repair\texorpdfstring{\\}{ }Without Per-State Expert Actions}

\author{%
\begin{tabular}{cc}
Peiying Zhu & Sidi Chang \\
Blossom AI & Blossom AI Labs \\
San Francisco, USA & Tokyo, Japan
\end{tabular}
}
\date{}

\hypersetup{
  pdftitle={When Aggregate Alignment Misleads: Auditing Policy Repair Without Per-State Expert Actions},
  pdfauthor={Peiying Zhu; Sidi Chang}
}

\begin{document}

\twocolumn[{
\begin{@twocolumnfalse}
\maketitle
\vspace{-1.6em}
\begin{abstract}
Agentic AI systems are increasingly used to edit, refine, and repair decision policies, but evaluating these edits is difficult when per-state expert action labels are unavailable. We study this problem in a hotel-pricing simulator where an agentic policy editor receives only region-level diagnostic feedback: summaries of how its price distribution differs from a benchmark policy across time, inventory, and market regions. The editor cannot observe benchmark actions, benchmark source code, reward numbers, or held-out outcomes, and may only propose constrained edits to a target-action table. On 5,000 held-out episodes, a multi-restart LLM editor reaches RevPAR 108.47 (95\% CI 107.61--109.34), close to the benchmark policy's 108.75 (107.81--109.68), with paired gap (LLM minus benchmark) -0.276 and 95\% CI [-0.692, 0.146]. A cheap diagnostic projection already recovers much of the revenue (107.90), so the LLM editor's distinctive gain is not raw revenue lift alone: it also reduces episode composition distance from 1.153 to 0.609. This is the strongest non-benchmark repair result. This profile is not explained by restart search alone: nonsemantic proposers with up to 2,500 evaluations fall 8.77--14.57 RevPAR points short. Nor is it explained by plausible prompt format: a shuffled-diagnostic control breaks region-error correspondence and falls to RevPAR 94.30. The match is genuine but partial. A tree editor achieves stronger pooled alignment, 0.214 versus 0.266, and stronger reference-state D1, 0.328 versus 1.197, yet revenue falls to 98.91. These results show that agentic policy repair should be evaluated by whether diagnostic feedback becomes reliable closed-loop outcome, not by a single behavioral distance.
\end{abstract}

\vspace{1em}
\end{@twocolumnfalse}
}]

\section{Introduction}

Agentic AI systems are increasingly asked to modify decision systems rather than merely answer questions. They write code, tune operational rules, update workflows, and propose policy edits. This creates a new evaluation problem. The agent's output is not a label or a prediction; it is an intervention on a decision policy. If the intervention looks plausible under a superficial metric, the system may still fail in the operational states that matter.

Recent work has emphasized language models as agents that reason, act, and interact with environments~\cite{yao2023react,liu2024agentbench}. Other benchmarks evaluate agents through artifact edits validated by external execution, such as code patches for software tasks~\cite{jimenez2024swebench}. Our setting follows the same broad direction but studies a different object: an agent edits a decision policy, and the evaluator must judge whether the policy repair is trustworthy without per-state expert action labels.

This paper studies agentic policy repair under missing per-state expert action labels. The agent is given a policy to edit and diagnostic feedback about how the policy differs from a benchmark. It is not given the benchmark action for each state, the benchmark implementation, reward numbers, or held-out outcomes. It must propose constrained edits to the policy. The evaluator then asks whether the edited policy is better.

This setting differs from standard supervised imitation and from reward optimization. In imitation learning, the learner observes expert state-action pairs. In reward optimization, the learner receives a scalar objective. In diagnostic-feedback repair, the agent receives aggregate behavioral summaries. These summaries may say that a region of the state space overuses or underuses certain actions, but they do not say what action to take in a particular state.

We use hotel pricing as a controlled testbed. The policy is a target-action table indexed by time, inventory, and market. The benchmark is a fixed rule-based revenue-management reference policy; it is a behavioral reference, not a claim of global optimality. The rule combines a myopic expected-revenue price, computed against a reference rival price, with an inventory-pace floor that raises the minimum acceptable price when sold inventory runs ahead of a target curve. The agent does not see this rule. It receives diagnostics comparing its price distribution with the benchmark across 12 operational regions. It can only make rectangular integer-shift edits to the target-action table. This setting gives us held-out outcomes, meaningful operational state variables, and an auditable edit grammar.

Why audit fidelity to a non-optimal benchmark? In many operational settings, a benchmark policy is a reference discipline rather than an optimality certificate. Operators may care that a repaired policy stays predictable, interpretable, and within the qualitative behavior of an approved rule, even when revenue is the ultimate business outcome. A revenue-only audit can accept a policy whose operational discipline has drifted. Our goal is therefore not to prove that benchmark imitation is always desirable, but to evaluate whether a repair preserves useful behavioral structure while improving or maintaining outcomes.

The key evaluation risk is aggregate alignment. A policy editor can make the overall action distribution resemble the benchmark while preserving little of the within-trajectory price composition that operators would recognize. In pricing, this means a policy may use the right global mix of prices but place that mix differently across booking episodes. More generally, a repaired policy may match a marginal action distribution while still failing a finer-grained behavioral audit.

We find that, in this controlled setting, region-level diagnostics are enough for an LLM editor to produce closed-loop value under a severe information restriction. The LLM editor reaches a RevPAR point estimate close to the benchmark, with a paired gap confidence interval that crosses zero, and achieves the best episode-composition distance among non-benchmark editors. A cheap diagnostic projection also reaches high revenue, so the LLM's distinctive contribution is not a large revenue lift over projection; it is the combination of near-benchmark revenue, stronger episode-composition fidelity, and controls showing that the region-level diagnostic correspondence matters. This is not merely restart search: compute-budget-matched nonsemantic proposers fall 8.77--14.57 RevPAR points short. It is not merely plausible prompt format: shuffled diagnostics collapse in revenue and episode-composition fidelity. The success is still partial. A tree-search editor achieves the best pooled action-distribution alignment and the best reference-state D1, but its held-out revenue collapses. Thus, in this simulator, reference-state behavioral fidelity is neither sufficient nor necessary for benchmark-level closed-loop revenue. The right evaluation question is whether diagnostic feedback becomes audited operational outcome.

We make three contributions. First, we formulate diagnostic-feedback policy repair as an agentic evaluation setting in which the agent edits a decision artifact without observing per-state benchmark actions, rewards, or held-out outcomes. Second, we show that a multi-restart LLM editor can translate localized diagnostic feedback into closed-loop revenue recovery under this restriction, while nonsemantic proposer controls and shuffled diagnostics cannot. Third, we show why this success still needs a multi-metric audit: behavioral alignment can look strong under one audit while revenue or another behavioral audit fails. The resulting protocol pairs outcome metrics, pooled alignment, episode-composition alignment, reference-state D1, strong projection baselines, nonsemantic proposer controls, and a shuffled-diagnostic placebo.

\section{Related Work}

LLM agents and feedback-driven refinement.
LLM agents combine language-model reasoning with actions in external environments~\cite{yao2023react}. Feedback-driven methods such as Reflexion and Self-Refine show that language models can improve outputs or decisions through iterative feedback without gradient updates~\cite{shinn2023reflexion,madaan2023selfrefine}. Our work shares the feedback-driven structure, but the feedback is neither free-form self-critique nor scalar reward. It is a region-level diagnostic comparing the edited policy's action distribution with a benchmark policy.

Agent evaluation.
AgentBench evaluates LLMs as agents across interactive environments~\cite{liu2024agentbench}, and recent surveys organize agent evaluation around objectives, processes, metrics, and deployment constraints~\cite{mohammadi2025agentsurvey}. SWE-bench studies language models that edit software artifacts and are evaluated by external tests~\cite{jimenez2024swebench}. Agentless further argues that constrained, interpretable repair pipelines can be competitive with more complex autonomous software agents~\cite{xia2024agentless}. We bring this evaluation lens to policy repair: the agent edits a decision policy, and success requires more than a single aggregate score.

Proxy objectives and specification gaming.
Our metric disagreement is related to reward hacking and specification gaming: optimizing a proxy can satisfy the written objective while missing the intended behavior~\cite{skalse2022rewardgaming}. Here the proxy is not the simulator reward itself, but a behavioral audit. The tree-search editor shows that a marginal action-distribution metric can be optimized while trajectory-local behavior and revenue fail.

Imitation and off-policy evaluation.
Inverse reinforcement learning and imitation learning study how to recover behavior from demonstrations or expert preferences~\cite{arora2021inverse}. Our setting removes per-state benchmark actions from the editor's input and exposes only region-level summaries. The audit problem is also adjacent to off-policy evaluation, where policy quality must be estimated under distributional and trajectory-level uncertainty~\cite{voloshin2021ope}. We use a simulator so that every selected policy can be audited on the same held-out production episodes.

\section{Diagnostic-Feedback Policy Repair}

We consider an agent that edits a decision policy. The editable policy is a target-action table. Each cell corresponds to a region of the operational state space, indexed by time, inventory, and market. The starting policy is a constant table. The benchmark policy is available only for diagnostic comparison and held-out audit.

At each repair iteration, the current policy is rolled out on training episodes. Its action distribution is compared with the benchmark policy's action distribution within each diagnostic region. The resulting diagnostic report summarizes how the current policy overuses or underuses action ranges in each region.

The agent receives this report and proposes a patch. A valid patch changes only target-action table entries through a rectangular integer shift. Invalid edits are rejected. The agent cannot modify the simulator, benchmark policy, reward function, metric definitions, or evaluator.

The information constraints are deliberate:

\begin{itemize}
    \item No per-state benchmark action labels are shown to the agent.
    \item No benchmark source code is shown to the agent.
    \item No reward numbers or held-out outcomes are shown in the prompt.
    \item No environment parameters are exposed beyond the diagnostic report.
    \item The final reported metrics are computed on held-out production episodes.
\end{itemize}

These constraints make the task an evaluation setting for diagnostic feedback. If the policy improves, the improvement must come from using diagnostic summaries to propose valid policy edits.

All reported policies are audited on the same 5,000 held-out production episodes. Candidate generation and selection use non-production diagnostics; production held-out outcomes are used once, after selection. This split is important because the paper's goal is not to optimize directly for production held-out revenue, but to evaluate whether diagnostic feedback can support policy repair under a locked decision space.

The protocol is designed around a specific evaluation threat model. Because an agent edits an artifact that will later be evaluated, a naive setup can confound policy repair with reward leakage, label leakage, or evaluator manipulation. Table~\ref{tab:threats} lists the main threats and the corresponding design choices in our study.

\begin{table*}[t]
\centering
\caption{Evaluation threat model for diagnostic-feedback policy repair.}
\label{tab:threats}
\small
\begin{tabular}{p{0.22\textwidth}p{0.34\textwidth}p{0.34\textwidth}}
\toprule
Threat & Why it matters & Design response \\
\midrule
Per-state action-label leakage & If the editor sees benchmark state-action labels, the task becomes imitation rather than diagnostic repair. & Prompts include only region-level distributional diagnostics, not per-state benchmark actions. \\
Reward leakage & If the editor sees reward numbers, it may optimize outcomes directly rather than use diagnostic feedback. & Prompts exclude reward, RevPAR, occupancy, ADR, and held-out outcomes. \\
Evaluator hacking & If the editor can change metrics, simulator state, or policy class, improvement may not reflect policy repair. & Patch validation permits only target-action table edits under the patch grammar. \\
Selector overfit & If selection and final audit use the same episodes, reported performance may reflect selection noise. & Candidate selection uses training diagnostics; final numbers use held-out production episodes once. \\
Prompt-format shortcut & If any plausible diagnostic prompt works, the diagnostic signal is not doing semantic work. & The shuffled-diagnostic placebo preserves format but breaks region-error correspondence. \\
\bottomrule
\end{tabular}
\end{table*}

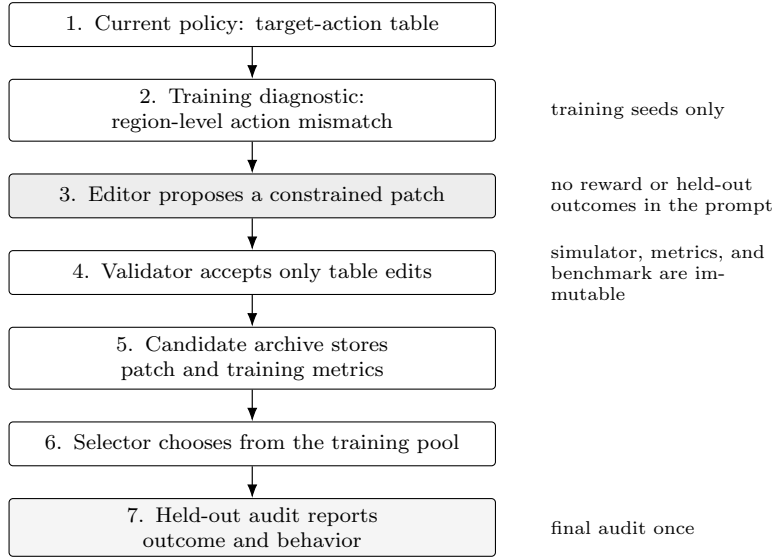
\begin{figure*}[t]
\centering
\begin{tikzpicture}[
    node distance=0.42cm and 0.6cm,
    box/.style={draw, rounded corners=1.5pt, align=center, text width=6.2cm, minimum height=0.58cm, font=\footnotesize},
    agent/.style={box, fill=black!7},
    audit/.style={box, fill=black!4},
    note/.style={align=left, font=\scriptsize, text width=3.2cm},
    arrow/.style={-{Latex[length=1.8mm]}, line width=0.45pt}
]
\node[box] (policy) {1. Current policy: target-action table};
\node[box, below=of policy] (diag) {2. Training diagnostic: region-level action mismatch};
\node[agent, below=of diag] (editor) {3. Editor proposes a constrained patch};
\node[box, below=of editor] (validator) {4. Validator accepts only table edits};
\node[box, below=of validator] (archive) {5. Candidate archive stores patch and training metrics};
\node[box, below=of archive] (selector) {6. Selector chooses from the training pool};
\node[audit, below=of selector] (audit) {7. Held-out audit reports outcome and behavior};
\draw[arrow] (policy) -- (diag);
\draw[arrow] (diag) -- (editor);
\draw[arrow] (editor) -- (validator);
\draw[arrow] (validator) -- (archive);
\draw[arrow] (archive) -- (selector);
\draw[arrow] (selector) -- (audit);
\node[note, right=of diag] {training seeds only};
\node[note, right=of editor] {no reward or held-out outcomes in the prompt};
\node[note, right=of validator] {simulator, metrics, and benchmark are immutable};
\node[note, right=of audit] {final audit once};
\end{tikzpicture}
\caption{Diagnostic-feedback policy repair pipeline. The editor observes region-level diagnostics and proposes constrained table edits; validation, candidate scoring, selection, and held-out auditing are external to the editor.}
\label{fig:pipeline}
\end{figure*}

Figure~\ref{fig:pipeline} summarizes the evaluation protocol. The key design choice is that the agent never owns the whole loop. It observes diagnostics and proposes patches, while separate components validate patches, score candidates, select among them, and audit the selected policy on held-out episodes.

\section{Policy Editors}

We compare three policy editors.

Diagnostic projection.
A direct projection maps diagnostic action distributions into the target-action table. This is a strong cheap baseline: it asks how much can be recovered by mechanically converting the diagnostic summary into a table, without language-model proposal generation. It is not a formal oracle because it does not search over all feasible tables.

Tree-search editor.
A nonsemantic tree-search editor searches over the same patch grammar using the same diagnostic schema. It tests whether the diagnostic representation and edit grammar are sufficient without language-model proposal structure. In the canonical run, tree search uses 500 candidate evaluations, beam width 5, branching factor 10, and a top-5 re-evaluation stage before the final production audit. Because this search budget is smaller than the LLM editor's 2,500 candidate evaluations, we use tree search primarily as a nonsemantic diagnostic consumer and metric stress test, not as a compute-matched proof that one proposal architecture dominates another.

This distinction matters for interpretation. The present paper evaluates whether a locked diagnostic-feedback repair protocol can be audited safely, not whether language-model proposals are intrinsically superior to every nonsemantic proposer. We therefore include additional up-to-2,500-evaluation random and diagnostic-guided nonsemantic proposer controls in Section~\ref{sec:robustness}.

Multi-restart LLM editor.
The LLM editor reads the diagnostic report and proposes one patch per iteration. The main run uses DeepSeek-V4-Flash at temperature 0 with JSON response mode and thinking disabled. It uses 125 restarts, 20 iterations per restart, and one candidate per iteration, yielding 2,500 candidate evaluations. Each restart begins from the same constant policy and uses a disjoint training seed partition. Candidate selection uses aggregate D1 plus Jensen-Shannon divergence on a 500-episode selector pool. Final metrics are reported only on held-out production episodes.

The LLM editor is therefore a constrained policy editor, not an autonomous pricing system. It proposes edits; external code validates, scores, selects, and audits them.

Table~\ref{tab:setup} locks the main implementation details for the LLM editor and the tree-search stress test.

\begin{table*}[t]
\centering
\caption{Locked implementation details for the main LLM editor and tree-search stress test.}
\label{tab:setup}
\small
\begin{tabular}{p{0.23\textwidth}p{0.67\textwidth}}
\toprule
Component & Setting \\
\midrule
Diagnostic schema & 12 regions from a $2\times2\times3$ partition over time, inventory, and market; each region reports action-distribution mismatch to the benchmark. \\
Editable policy & Target-action table with shape $30\times30\times10$ over time, inventory, and market. The starting policy is a constant target-action table. \\
Patch grammar & JSON patches with at most 16 updates. Allowed operations are \texttt{shift\_region} and \texttt{set\_region} over half-open rectangular ranges. Simulator parameters, market edges, bin edges, $\tau$, metrics, and policy class are immutable. \\
LLM proposer & DeepSeek-V4-Flash, temperature 0, JSON response format, maximum 2,048 output tokens. All 2,500 returned patches were accepted by the validator. \\
LLM search and selection & 125 restarts $\times$ 20 iterations $\times$ 1 candidate. Each restart uses a disjoint 100-seed training partition. Restart-top candidates are re-scored on a 500-seed selector pool using aggregate D1 plus Jensen-Shannon divergence. \\
Tree-search stress test & 500 candidate evaluations, beam width 5, branching factor 10, same diagnostic schema and patch grammar, top-5 re-evaluation before production audit. \\
Final audit & 5,000 held-out production episodes, used only after a policy is selected. \\
\bottomrule
\end{tabular}
\end{table*}

For reproducibility, the LLM prompt is structured in four blocks. The instruction block names the permitted JSON keys and patch operations. The policy-summary block reports the policy class, table shape, action histogram, and region-block modes. The restart block reports only the restart identifier. The diagnostic block reports the 12-region schema, aggregate D1 and Jensen-Shannon values, and top region records containing support counts, current and reference action histograms, mean-action gaps, and recommended shift directions. The prompt does not include reward, RevPAR, occupancy, ADR, held-out outcomes, benchmark source code, or per-state benchmark actions.\ The run artifact stores the rendered prompt, raw response, parsed patch, validator result, token usage, response model, and system fingerprint for every candidate. DeepSeek's API did not receive a seed parameter; the run uses temperature 0 and fixed prompts, but service-side nondeterminism remains possible.

Figure~\ref{fig:algorithm} gives the multi-restart procedure used for the LLM editor. The pseudo-code is included because the restart structure is part of the evaluation design: each restart begins from the same starting policy, uses a disjoint training partition, and contributes a restart-best candidate to the final selector pool.

\begin{figure}[t]
\centering
\small
\setlength{\tabcolsep}{3pt}
\begin{tabular}{@{}r p{0.82\columnwidth}@{}}
1 & Input starting policy $\pi_0$, diagnostic generator $G$, patch grammar $\mathcal{P}$, restarts $M=125$, iterations $L=20$. \\
2 & For each restart $m=1,\ldots,M$: \\
3 & \quad Set $\pi \leftarrow \pi_0$. \\
4 & \quad For each iteration $\ell=1,\ldots,L$: \\
5 & \quad\quad Roll out $\pi$ on restart training seeds. \\
6 & \quad\quad Compute diagnostic $d \leftarrow G(\pi)$. \\
7 & \quad\quad Ask editor for patch $p \in \mathcal{P}$. \\
8 & \quad\quad Reject invalid patches; otherwise apply $p$. \\
9 & \quad\quad Score candidate on restart diagnostics. \\
10 & \quad Keep restart-best candidate. \\
11 & Re-score restart-best candidates on selector pool. \\
12 & Select one policy and audit once on held-out episodes. \\
\end{tabular}
\caption{Algorithmic view of the multi-restart diagnostic policy editor. The LLM proposes patches, but validation, scoring, selection, and final audit are external.}
\label{fig:algorithm}
\end{figure}

\section{Evaluation Metrics}

We evaluate policies using operational and behavioral metrics.

RevPAR, occupancy, and ADR measure operational performance in the hotel-pricing simulator. In each episode, RevPAR is gross revenue divided by available rooms, and occupancy is sold rooms divided by available rooms. Tables report mean RevPAR and mean occupancy over held-out episodes. ADR is reported as aggregate revenue per sold room over the same held-out episode set, equivalently mean RevPAR divided by mean occupancy; therefore RevPAR equals ADR times occupancy up to rounding.

D1 distance is the L1 distance between two discrete action histograms: $D_1(p,q)=\sum_a |p(a)-q(a)|$. Since both histograms sum to one, D1 lies in $[0,2]$, with 0 meaning identical action distributions.

Pooled D1 measures action-distribution distance after pooling all held-out episodes from each policy's own rollout distribution. It asks whether the edited policy uses the same global action mix as the benchmark. If $P_{\pi}^{pool}$ and $P_{b}^{pool}$ are the pooled action distributions of policy $\pi$ and benchmark $b$, pooled D1 is $D_1(P_{\pi}^{pool}, P_b^{pool})$.

Episode D1 computes action-distribution distance within each held-out episode and averages across episodes, again using each policy's own closed-loop rollout states. For held-out episodes $e=1,\ldots,N$, it is $N^{-1}\sum_e D_1(P_{\pi}^{e}, P_b^{e})$. This is still a marginal histogram metric: it does not compare the action chosen at each state. Its value is that the marginal is computed at a finer unit, so each held-out trajectory must have a price composition closer to the benchmark's trajectory-level composition.

Reference-state D1 is a counterfactual behavioral stress test, abbreviated as Ref-state D1 in tables. We replay the benchmark on held-out seeds, group the benchmark's visited states into the same 12 diagnostic regions, and evaluate the edited policy's action distribution on those benchmark states. For region $r$, we compare $P_{\pi}(a\mid r, s\sim d_b)$ with $P_b(a\mid r, s\sim d_b)$ and average by benchmark region visitation share. This metric answers a different question from pooled or episode D1: how the edited policy would act on benchmark-visited states. Because it evaluates under the benchmark state distribution rather than the edited policy's own closed-loop distribution, we use it as a stress test rather than as the primary success metric.

This distinction is the central audit design. Pooled D1 measures global marginal alignment under each policy's own closed-loop state distribution. Episode D1 measures trajectory-local marginal alignment under those same policy-specific rollouts. Reference-state D1 evaluates action composition on benchmark-visited states, $s\sim d_b$. A policy can look aligned under one of these metrics while failing another because the metrics condition on different units and, for reference-state D1, a different state distribution.

The behavioral metrics are intentionally complementary. Pooled D1 is useful because it catches whether a repair moved the overall action mix toward the benchmark. Episode D1 is useful because the global action mix may be correct while individual episode compositions remain different. Reference-state D1 is useful as a counterfactual stress test, but it should be interpreted with outcome and closed-loop episode metrics. No single one is sufficient for closed-loop policy certification. The tree-search result in Section~\ref{sec:results} is a concrete example of this disagreement.

Reported confidence intervals are 95\% intervals over the 5,000 held-out production episodes for operational and episode-level metrics. Pooled and reference-state D1 intervals are computed by bootstrap resampling held-out episodes. Because all policies are evaluated on the same held-out seeds, paired resampling is the sharper way to compare two selected policies directly; we use it for the main RevPAR contrast between the LLM editor and the benchmark.

\section{Held-Out Results}
\label{sec:results}

Table~\ref{tab:main} reports held-out evaluation over 5,000 production episodes.

\begin{table*}[t]
\centering
\caption{Held-out evaluation over 5,000 production episodes.}
\label{tab:main}
\begin{tabular}{lrrrrrr}
\toprule
Policy / editor & RevPAR & Occupancy & ADR & Pooled D1 & Episode D1 & Ref-state D1 \\
\midrule
Starting policy & 102.751 & 0.8563 & 120.00 & 1.153 & 1.153 & 1.153 \\
Benchmark policy & 108.749 & 0.7696 & 141.31 & 0.000 & 0.000 & 0.000 \\
Diagnostic projection & 107.902 & 0.8020 & 134.54 & 0.485 & 0.685 & 1.108 \\
Tree-search editor & 98.911 & 0.7582 & 130.45 & 0.214 & 0.766 & 0.328 \\
Multi-restart LLM editor & 108.472 & 0.7963 & 136.21 & 0.266 & 0.609 & 1.197 \\
\bottomrule
\end{tabular}
\end{table*}

The LLM editor recovers a RevPAR point estimate close to the benchmark on this held-out split. It raises RevPAR from 102.751 to 108.472, close to the benchmark's 108.749. It also reduces episode-composition distance from 1.153 to 0.609, the strongest non-benchmark result. It does not win every behavioral audit: its reference-state D1 is 1.197, worse than both the starting policy's 1.153 and the diagnostic projection's 1.108. We treat this as stress-test evidence, not as the primary success criterion.

The confidence intervals sharpen the interpretation. The LLM editor's RevPAR interval, 107.61--109.34, and the benchmark interval, 107.81--109.68, are close at this audit size. A paired bootstrap over the same 5,000 held-out seeds estimates the LLM-minus-benchmark RevPAR gap as -0.276 with 95\% CI [-0.692, 0.146]. We therefore describe the revenue result as close to the benchmark, not as exact equality. The behavioral contrasts move in different directions: LLM episode D1 is 0.609 (0.599--0.618) versus diagnostic projection episode D1 0.685 (0.675--0.695) and tree-search episode D1 0.766 (0.756--0.776), while tree-search pooled D1 is 0.214 (0.214--0.215) versus LLM pooled D1 approximately 0.266--0.267. Under reference-state D1, tree search is much closer to the benchmark, 0.328 (0.322--0.340), than the LLM editor, 1.197 (1.188--1.207).

The tree-search editor exposes the evaluation failure. It achieves pooled D1 of 0.214 and reference-state D1 of 0.328, the strongest behavioral alignment under those two audits. If either metric were used alone, tree search would appear strongest. But its RevPAR falls to 98.911 and its episode D1 is 0.766. It matches important behavioral summaries while failing on revenue and trajectory-local price composition.

The diagnostic projection is a strong baseline. It reaches RevPAR 107.902, only 0.570 below the LLM editor, which means the main value of the LLM editor is not a large revenue lift over projection in this simulator. The LLM improves pooled D1 from 0.485 to 0.266 and episode D1 from 0.685 to 0.609, but worsens reference-state D1 from 1.108 to 1.197. This suggests that revenue recovery, marginal behavioral recovery, and counterfactual reference-state behavior are related but not identical.

As a robustness check, we also recomputed the distributional comparisons using Wasserstein-1 distance, or Earth Mover's Distance, in dollars. The qualitative conclusions do not change. Pooled W1 still ranks tree search ahead of the LLM editor (3.66 versus 6.05), episode W1 still ranks the LLM editor best among non-benchmark policies (10.37 versus 11.23 for projection and 15.18 for tree search), and reference-state W1 still ranks tree search ahead of the LLM editor (6.04 versus 16.69). Thus the central metric disagreement is not an artifact of categorical L1 distance.

The main result is therefore not simply that the LLM editor improves the policy. The stronger result is that the LLM editor turns localized diagnostic feedback into closed-loop outcome under a locked edit space, while single behavioral metrics can misrank repairs. The evaluation target should not be which behavioral distance is smallest in isolation. It should be whether a diagnostic-driven edit remains useful under held-out closed-loop execution.

Table~\ref{tab:metricfailure} shows the same point as a single-metric audit failure analysis. Each metric answers a useful question, but each can also support an incomplete conclusion if used alone.

\begin{table*}[t]
\centering
\caption{Why no single metric is sufficient for policy-repair evaluation.}
\label{tab:metricfailure}
\small
\begin{tabular}{p{0.19\textwidth}p{0.24\textwidth}p{0.27\textwidth}p{0.20\textwidth}}
\toprule
Metric used alone & Apparent conclusion & What it misses & Evidence in our audit \\
\midrule
RevPAR & Projection and LLM editor both recover high revenue. & Revenue recovery does not imply behavioral recovery. & Projection reaches 107.902 RevPAR but episode D1 remains 0.685. \\
Pooled D1 & Tree-search editor looks best aligned. & Aggregate action mix can hide trajectory-local and revenue failures. & Tree pooled D1 is 0.214 but RevPAR falls to 98.911. \\
Episode D1 & LLM editor is the strongest non-benchmark repair. & Repair is still partial and does not imply success under every stress test. & LLM episode D1 is 0.609 but reference-state D1 is 1.197. \\
Reference-state D1 & Tree-search editor looks best aligned. & Counterfactual reference-state alignment can still fail as a closed-loop policy. & Tree reference-state D1 is 0.328 but RevPAR falls to 98.911. \\
Shuffled pooled D1 & Placebo appears close to true diagnostic. & Prompt format and global mix do not imply correct region-error use. & Shuffled pooled D1 is 0.294 but RevPAR is 94.30. \\
\bottomrule
\end{tabular}
\end{table*}

\section{Shuffled-Diagnostic Placebo}

We next test whether the diagnostic feedback matters because it links errors to the correct regions. The shuffled-diagnostic control preserves the prompt format, patch grammar, and selection machinery, but shuffles diagnostic contents across region coordinates. A mismatch summary from one region can appear under another region. The report still looks like a valid diagnostic report, but the operational mapping is wrong.

Scoring and selection still use true diagnostics. This makes the control conservative: a candidate produced from shuffled feedback can still be selected if it accidentally scores well under the true diagnostic objective.

This control is a mechanism placebo rather than a compute-matched leaderboard run. It uses 10 restarts and 200 LLM calls, compared with 125 restarts and 2,500 calls in the main run. We use it to test whether a plausible diagnostic prompt without correct region-error correspondence is sufficient, not to estimate the best possible shuffled-diagnostic policy under the full budget. Table~\ref{tab:placebo} reports the resulting held-out audit.

\begin{table*}[t]
\centering
\caption{Shuffled-diagnostic control.}
\label{tab:placebo}
\scriptsize
\setlength{\tabcolsep}{2.5pt}
\begin{tabular}{lrrrrrr}
\toprule
Condition & RevPAR & Occ. & ADR & Pool D1 & Epis. D1 & Ref-state D1 \\
\midrule
True diagnostic & 108.47 & 0.7963 & 136.21 & 0.266 & 0.609 & 1.197 \\
Shuffled selector-best & 94.30 & 0.7252 & 130.03 & 0.294 & 0.919 & 1.485 \\
Shuffled heldout-best & 102.60 & 0.7956 & 128.97 & 0.456 & 0.873 & 1.316 \\
\bottomrule
\end{tabular}
\end{table*}

The shuffled selector-best policy obtains pooled D1 of 0.294, close to the true diagnostic editor's 0.266. Yet RevPAR collapses to 94.30 and episode D1 rises to 0.919. The heldout-best shuffled policy among shards reaches RevPAR 102.60, roughly the starting policy, but still has weak episode-composition fidelity.

The selector-best shuffled RevPAR interval is 93.46--95.14 and its episode D1 interval is 0.909--0.929, far from the main LLM editor. The heldout-best shuffled row is post-hoc optimistic: it selects the best held-out RevPAR among the five shuffled shards after audit and is included only to show that even the most favorable shuffled shard does not recover the main result.

This control shows why diagnostic localization matters. The agent is not merely benefiting from a plausible global action histogram. The diagnostic must connect behavioral errors to the correct operational regions. When that connection is broken, pooled alignment can remain deceptively good while operational performance and episode-composition fidelity fail.

\section{Robustness and Audit Checks}
\label{sec:robustness}

Restart budget.
The LLM editor improves as the restart budget increases, as shown in Table~\ref{tab:restart}. The best policy appears only after enough independent restarts, supporting the use of a multi-restart editor rather than a single repair trajectory.

\begin{table*}[t]
\centering
\caption{Restart budget sensitivity for the LLM editor.}
\label{tab:restart}
\scriptsize
\setlength{\tabcolsep}{5pt}
\begin{tabular}{lrr}
\toprule
Restart budget & RevPAR & Episode D1 \\
\midrule
5 restarts & 105.50 & 0.647 \\
25 to 75 restarts & 106.62 & 0.618 \\
100 to 125 restarts & 108.47 & 0.609 \\
\bottomrule
\end{tabular}
\end{table*}

Selector sensitivity.
Several selector variants choose the same final candidate. The main behavioral selector, aggregate D1 plus Jensen-Shannon divergence, selects the same candidate as D1 alone and Jensen-Shannon divergence alone. This creates a useful tension: the selector is still a marginal behavioral objective, while the paper argues that marginal objectives are incomplete. We handle this by treating selection as only one stage of the protocol and requiring final held-out audits with outcome, pooled, episode-composition, and reference-state D1. We also ran reward-aware selector variants as post-hoc sensitivity checks; these are not part of the policy-selection protocol, but they selected the same candidate in this archive. This reduces concern that the main result is an artifact of one fragile scalarization while keeping the reported selection reward-blind.

Compute-matched nonsemantic proposers.
To test whether the result is merely due to restart search rather than language-model proposal structure, we ran two nonsemantic tree proposers with the same diagnostic schema, patch grammar, selector objective, and an up-to-2,500 candidate-evaluation budget. A random-region proposer reached RevPAR 93.90 and episode D1 0.883. A diagnostic-guided region proposer improved to RevPAR 99.70 and episode D1 0.746, but remained well below the LLM editor's RevPAR 108.47 and episode D1 0.609. Their pooled D1 and reference-state D1 values can be lower than the LLM editor's, which reinforces the central audit point: behavioral distances can look excellent while closed-loop revenue and trajectory-local composition fail. Table~\ref{tab:nonsemantic} summarizes the comparison.

\begin{table*}[t]
\centering
\caption{Up-to-2,500-evaluation nonsemantic proposer controls.}
\label{tab:nonsemantic}
\scriptsize
\setlength{\tabcolsep}{2pt}
\begin{tabular}{lrrrrrrr}
\toprule
Proposer & Evals & RevPAR & Occ. & ADR & Pool D1 & Epis. D1 & Ref-state D1 \\
\midrule
Random region & 2461 & 93.90 & 0.737 & 127.41 & 0.090 & 0.883 & 0.680 \\
Guided region & 2461 & 99.70 & 0.751 & 132.67 & 0.074 & 0.746 & 0.422 \\
LLM editor & 2500 & 108.47 & 0.796 & 136.21 & 0.266 & 0.609 & 1.197 \\
\bottomrule
\end{tabular}
\vspace{2pt}
\begin{minipage}{0.96\textwidth}
\scriptsize Evals are completed candidate evaluations under a 2,500 cap; the tree schedule stops at 2,461 because of the fixed beam-depth schedule.
\end{minipage}
\end{table*}

Candidate diversity.
Across 2,500 candidates, the run produced 480 unique policies. Among the 125 restart-top policies, 69 were unique. This suggests that the editor did not simply repeat one patch across all restarts.

Decision-space lock.
Prompt construction excludes benchmark state-action labels, benchmark source code, reward numbers, held-out outcomes, and environment parameters. Patch application only modifies target-action table entries.

Together, these checks support a diagnostic-feedback interpretation rather than reward leakage, benchmark-label leakage, or evaluator hacking.

Table~\ref{tab:evidence} summarizes what each experiment contributes to the paper's evaluation claim. The table is included to make clear that the evidence is not a single leaderboard row: the main result, negative control, placebo, and audit checks each rule out a different failure explanation.

\begin{table*}[t]
\centering
\caption{Evidence map for the evaluation protocol.}
\label{tab:evidence}
\small
\begin{tabular}{p{0.20\textwidth}p{0.32\textwidth}p{0.38\textwidth}}
\toprule
Experiment / check & What it tests & Conclusion supported \\
\midrule
Main held-out audit & Whether the selected repair improves operational and behavioral metrics on unseen episodes & LLM editor reaches a paired RevPAR gap CI that crosses zero and best episode D1 among non-benchmark policies, but not best reference-state D1. \\
Tree-search editor & Whether behavioral alignment alone certifies a good repair & No. Tree has strongest pooled and reference-state D1 but loses revenue and has worse episode D1. \\
Diagnostic projection & Whether direct projection recovers both revenue and behavior & It recovers much of the revenue profile but not benchmark-like behavior. \\
Nonsemantic proposer controls & Whether up-to-2,500 random or guided patch search recovers the LLM result & No. The best nonsemantic proposer reaches RevPAR 99.70 and episode D1 0.746, despite pooled D1 0.074 and reference-state D1 0.422. \\
Shuffled diagnostic & Whether prompt format and global action mix are enough & No. Shuffling preserves similar pooled alignment but breaks revenue and episode-composition fidelity. \\
Restart sensitivity & Whether the final policy is available in a very small search budget & No. Strongest policy appears only after enough independent restarts. \\
Selector sensitivity & Whether the main result depends on one fragile selector & No. Multiple selector variants choose the same final candidate. \\
Decision-space lock & Whether the agent can alter the evaluator or policy class & No. Patches only modify target-action table entries. \\
\bottomrule
\end{tabular}
\end{table*}

\section{Discussion}

This paper studies policy repair as an agentic evaluation problem. The editor is not merely producing text or predictions. It is proposing interventions on a decision policy. Such interventions can look good under one metric and fail under another.

The tree-search result illustrates the danger. Tree search optimizes toward behavioral similarity and obtains the best pooled alignment in the main table, but the resulting policy loses revenue. Reference-state D1 makes the point sharper rather than weaker: a counterfactual behavioral metric can rank tree search ahead while the closed-loop policy is operationally poor. The LLM result shows the complementary profile: near-benchmark revenue and improved episode composition, but weak reference-state behavior. In this simulator, reference-state behavioral fidelity is not sufficient for good closed-loop revenue, because tree search has strong reference-state D1 but fails operationally. It is also not necessary, because the LLM editor has weak reference-state D1 but reaches near-benchmark revenue. This does not make the stress test useless. It makes it one audit rather than a leaderboard. A behavioral proxy may capture one aspect of the benchmark while missing trajectory composition or outcome dynamics; an outcome metric may capture revenue while missing behavioral drift.

One plausible mechanism is that the diagnostic available to the editor is regional and distributional rather than per-state. An LLM patch can improve trajectory-level price composition by shifting price mass across table regions, even if it does not reproduce how the benchmark would act on the benchmark's own visited states. This is why the LLM can improve episode D1 while worsening reference-state D1. If temporal and operational behavior matter, aggregate alignment is not enough.

The shuffled-diagnostic control provides a useful evaluation template. It preserves surface form while breaking the semantic mapping that should matter. If an agent performs similarly under true and shuffled diagnostics, the claimed diagnostic signal is probably not driving the result. In our experiment, shuffled diagnostics preserve much of the pooled alignment but fail on revenue and episode-composition fidelity.

These findings suggest a broader recipe for evaluating agentic policy repair, summarized in Table~\ref{tab:audit}.

\begin{table}[t]
\centering
\caption{Audit checks for agentic policy repair.}
\label{tab:audit}
\scriptsize
\setlength{\tabcolsep}{3pt}
\begin{tabular}{p{0.47\columnwidth}p{0.43\columnwidth}}
\toprule
Failure mode & Audit check \\
\midrule
Revenue improves but behavior drifts & Episode D1 + reference-state D1 \\
Global mix hides trajectory drift & Episode D1 + outcome \\
Reference-state proxy has false positives & Reference-state D1 interpreted with outcome \\
Agent ignores diagnostic localization & Shuffled diagnostic \\
Agent exploits edit space & Patch validator \\
Selector overfits diagnostics & Held-out audit \\
\bottomrule
\end{tabular}
\end{table}

\section{Engineering Interpretation}

The proposed protocol is best understood as an offline policy-repair assistant, not as an autonomous deployed decision maker. In an engineering deployment, the agent would sit between a diagnostic generator and an offline evaluator. It would propose candidate patches, but the production system would enforce the patch grammar, reject invalid edits, evaluate candidates on replay or simulation, and require a held-out audit before any shadow or canary deployment.

This separation of responsibilities is the practical value of the design. The LLM is used for semantic proposal generation, where it can translate diagnostic language into structured edits. It is not trusted with reward computation, metric definitions, environment parameters, or final acceptance. This makes the agent auditable and reduces the risk that it improves a superficial signal by changing the wrong part of the system.

The experiment also suggests what an engineering dashboard should include. A dashboard that reports only reward can miss behavior drift. A dashboard that reports only pooled action-distribution distance can miss trajectory-local failures. A safer dashboard reports outcome metrics, pooled alignment, episode-composition alignment, reference-state D1, diagnostic placebo performance, patch validity, and the training-versus-heldout gap. The reference-state metric should still be treated as one audit, not as a replacement for closed-loop held-out evaluation.

\section{Limitations}

The study uses one simulator and one benchmark policy rather than multiple domains. The benchmark policy is a reference policy, not a universal optimum. The main LLM result is one production run; because the DeepSeek service does not expose a seed parameter, run-level variance remains to be measured. The LLM editor still has substantial residual episode-level and reference-state distance from the benchmark. Episode D1 is stricter than pooled D1 but is still a trajectory-local marginal metric rather than exact per-state imitation. Reference-state D1 is a counterfactual stress test under the benchmark state distribution; it is useful for audit disagreement but is not a primary closed-loop success metric. The shuffled-diagnostic control is smaller than the main LLM run. The nonsemantic proposer controls reduce the concern that restart search alone explains the LLM result, but they do not exhaust all possible handcrafted heuristics. Finally, the target-action table is one policy representation; other policy classes may require different patch grammars and audit metrics.

These limitations define the scope of the paper. We do not claim that LLM agents generally repair decision policies. We show that diagnostic-feedback repair can be evaluated with stronger audits, and that aggregate alignment can be misleading in a controlled policy-repair setting.

\section{Conclusion}

Agentic policy editors need evaluation protocols that go beyond aggregate alignment. In a hotel-pricing simulator, a multi-restart LLM editor converts region-level diagnostic feedback into near-benchmark closed-loop revenue without per-state benchmark action labels, reward numbers, or held-out outcomes. Compute-budget-matched nonsemantic proposers and shuffled diagnostics do not recover this result, which supports the claim that the LLM is using diagnostic feedback rather than merely benefiting from restart search or prompt plausibility. The success is partial: the LLM editor improves episode-composition fidelity but does not win every behavioral stress test. Conversely, nonsemantic proposers can achieve excellent pooled or reference-state alignment while losing revenue. These results show that trustworthy agentic policy repair requires multi-granularity behavioral audits and diagnostic placebo controls. The question is not only whether the edited policy matches a behavioral summary. It is whether diagnostic feedback becomes reliable closed-loop outcome.

\appendix

\section{Reproducibility Details}

This appendix lists the information needed to reproduce the reported experiment from the repository. All commands should be run from the repository root at commit \texttt{5c2d8b0}. In the notes below, \texttt{PROJECT} denotes \path{projects/hotel-pricing-marl}. Python dependencies are resolved with \texttt{uv run} using \texttt{PROJECT} as the project path. For a fresh reproduction, write artifacts under \texttt{RUN\_ROOT}=\path{PROJECT/runs/policy_repair/}, with subfolders \path{llm_main/}, \path{diagnostic_placebo/}, \path{nonsemantic_controls/}, and \path{metrics/}. The LLM run requires \path{DEEPSEEK_API_KEY} in the environment. The DeepSeek wrapper uses model \texttt{deepseek-v4-flash}, JSON response mode, \texttt{temperature=0.0}, \texttt{max\_tokens=2048}, and does not send a seed parameter because the service does not expose one in the chat-completions API.

\subsection{Frozen Data Splits and Edit Space}

The training rollout pool is \texttt{100000..119999}. The production held-out pool is \texttt{200000..204999}. The smoke held-out pool is \texttt{300000..300499}. For the main multi-restart LLM editor, the selector pool is the first 500 training seeds, the next 500 training seeds are unused buffer seeds, and restart \texttt{i} uses training seeds \texttt{101000+100i} through \texttt{101099+100i}. The initial policy is the constant target-action policy with action bucket 1. The only mutable object is \texttt{target\_actions}; the patch validator rejects edits to the benchmark policy, simulator, metrics, bin edges, or evaluation pipeline.

\subsection{Main LLM Editor Run}

The main run uses \path{PROJECT/scripts/run_cp14_tqm_llm_cp13_market.py}. The key parameter values are: run mode \texttt{v19\_multi\_restart}, schema \texttt{S3}, granularity \texttt{dslm\_2x2x3}, objective \texttt{aggregate}, 125 restarts, 20 iterations per restart, initial target action 1, production held-out pool, and 5,000 held-out seeds. The proposer wrapper is \path{PROJECT/scripts/cp14_deepseek_proposer.py} with model \texttt{deepseek-v4-flash} and max output tokens 2048. The product of restarts and iterations gives 2,500 candidate policy evaluations. The runner writes all accepted patches, rejected/no-op patches, candidate metrics, restart ids, selected policy, and held-out audit metrics under \path{RUN_ROOT/llm_main/}.

\subsection{Controls and Table Regeneration}

The shuffled-diagnostic placebo uses the same runner and settings as the main run, with diagnostic-control mode \texttt{shuffled\_regions}. In our run it was sharded with \path{PROJECT/scripts/run_cp14_tqm_placebo_shard_20260524.sh}; each shard uses two restarts, and five shards give 200 LLM calls. Placebo artifacts should be grouped under \path{RUN_ROOT/diagnostic_placebo/}. The random and diagnostic-guided nonsemantic controls use \path{PROJECT/scripts/run_cp14_tqm_tree_cp13_market.py} with the same schema, initial policy, held-out pool, and 2,500 evaluation budget. The two proposal modes are \texttt{random\_region} and \texttt{tqm\_aligned\_guided\_region}; their artifacts should be grouped under \path{RUN_ROOT/nonsemantic_controls/}. To regenerate the context-conditional and reference-state metrics used in the paper, run \path{PROJECT/scripts/compute_kdd_state_conditional_metrics.py}; it replays the archived policies on the production held-out seeds and writes both JSON and CSV summaries under \path{RUN_ROOT/metrics/}.

\bibliographystyle{ACM-Reference-Format}
\bibliography{references}

\end{document}